\documentclass{article}

\usepackage{microtype}
\usepackage{graphicx}
\usepackage{subfigure}
\usepackage{booktabs} 
\usepackage{graphicx}
\usepackage{stackengine}
\usepackage{array}
\usepackage{subfigure}
\usepackage{tabularx}
\usepackage{multirow}
\usepackage{caption}
\usepackage{algpseudocode}
\usepackage{amssymb}
\usepackage{xr}
\usepackage{enumitem}
\usepackage{pdfpages}

\usepackage{color}

\usepackage[english]{babel}

\usepackage{color}
\usepackage{bm}
\usepackage{hyperref}
\usepackage{amsmath,amsfonts,amsthm}
\usepackage{blindtext}
\usepackage{listings}

\newif\ifdebug
\debugfalse

\ifdebug
\newcommand{\chris}[1]{{\color{red}{\bf\sf [CJF: #1]}}}
\newcommand{\junz}[1]{{\color{blue}{\bf\sf [ZJ: #1]}}}
\newcommand{\lkw}[1]{{\color{cyan}{\bf\sf [LKW: #1]}}}
\newcommand{\cwg}[1]{{\color{green}{\bf\sf [CWG: #1]}}}
\newcommand{\cx}[1]{{\color{green}{\bf\sf [CX: #1]}}}
\else
\newcommand{\chris}[1]{{\color{red}{}}}
\newcommand{\junz}[1]{{\color{blue}{}}}
\newcommand{\lkw}[1]{{\color{cyan}{}}}
\newcommand{\cwg}[1]{{\color{green}{}}}
\newcommand{\cx}[1]{{\color{green}{}}}
\fi

\newcommand{\vect}[1]{\boldsymbol{\mathbf{#1}}}

\newcommand{\E}{\mathbb{E}}

\newcommand{\nv}{\vect n}

\newcommand{\rv}{\vect r}

\newcommand{\Bc}{\mathcal B}

\newcommand{\Ec}{\mathcal E}

\newcommand{\Lc}{\mathcal L}

\newcommand{\Uc}{\mathcal U}
\newcommand{\Vc}{\mathcal V}

\newcommand{\norm}[1]{\left\lVert#1\right\rVert}
\newcommand{\abs}[1]{\left|#1\right|}
\newcommand{\inorm}[1]{\norm{#1}_\infty}
\newcommand{\iprod}[1]{\langle#1\rangle}

\newcommand{\Var}{\mbox{Var}}
\newcommand{\Cov}{\mbox{Cov}}

\newcommand{\bl}{{(l)}}
\newcommand{\bll}{{(l+1)}}
\newcommand{\bL}{{(L)}}
\newcommand{\NS}{\mbox{NS}}
\newcommand{\CV}{\mbox{CV}}
\newcommand{\CVD}{\mbox{CVD}}

\newcommand{\nbr}{\nv(u)}
\newcommand{\snbr}{\hat \nv^\bl(u)}
\newcommand{\var}[2]{\Var_{#1}\left[#2\right]}

\newcommand{\rh}{\mathring{h}}

\makeatletter
\newtheorem*{rep@theorem}{\rep@title}
\newcommand{\newreptheorem}[2]{%
	\newenvironment{rep#1}[1]{%
		\def\rep@title{#2 \ref{##1}}%
		\begin{rep@theorem}}%
		{\end{rep@theorem}}}
\makeatother

\newtheorem{proposition}{Proposition}
\newtheorem{theorem}{Theorem}

\newreptheorem{proposition}{Proposition}
\newreptheorem{theorem}{Theorem}
\newreptheorem{lemma}{Lemma}

\usepackage{hyperref}

\usepackage[accepted]{icml2018}

\icmltitlerunning{Stochastic Training of Graph Convolutional Networks with Variance Reduction}

\begin{document}

\twocolumn[
\icmltitle{Stochastic Training of Graph Convolutional Networks with Variance Reduction}



\icmlsetsymbol{equal}{*}

\begin{icmlauthorlist}
\icmlauthor{Jianfei Chen}{tsinghua}
\icmlauthor{Jun Zhu}{tsinghua}
\icmlauthor{Le Song}{getech,ant}
\end{icmlauthorlist}

\icmlaffiliation{tsinghua}{Dept. of Comp. Sci. \& Tech., TNList Lab, State Key Lab for Intell. Tech. \& Sys., Tsinghua University, Beijing, 100084, China}
\icmlaffiliation{getech}{Georgia Institute of Technology}
\icmlaffiliation{ant}{Ant Financial}

\icmlcorrespondingauthor{Jun Zhu}{dcszj@mail.tsinghua.edu.cn}

\icmlkeywords{Machine Learning, ICML}

\vskip 0.3in
]



\printAffiliationsAndNotice{}  

\begin{abstract}
Graph convolutional networks (GCNs) are powerful deep neural networks for graph-structured data. However, GCN computes the representation of a node recursively from its neighbors, making the receptive field size grow exponentially with the number of layers. Previous attempts on reducing the receptive field size by subsampling neighbors do not have a convergence guarantee, and their receptive field size per node is still in the order of hundreds.
In this paper, we develop control variate based algorithms which allow sampling an arbitrarily small neighbor size. Furthermore, we prove new theoretical guarantee for our algorithms to converge to a local optimum of GCN. Empirical results show that our algorithms enjoy a similar convergence with the exact algorithm using only two neighbors per node. The runtime of our algorithms on a large Reddit dataset is only one seventh of previous neighbor sampling algorithms.
\end{abstract}

\newcommand{\propositionvar}{
If $\hat \nv^\bl(u)$ contains $D^\bl$ samples from  $\nv(u)$ without replacement, then
$\Var_{\hat \nv^\bl(u)}\left[\frac{n(u)}{D^\bl}\sum_{v\in \hat \nv^\bl(u)} x_v \right] 
= \frac{C_u^\bl}{2D^\bl} \sum_{v_1\in \nv(u)}\sum_{v_2\in\nv(u)} (x_{v_1}-x_{v_2})^2$, where $C_u^\bl = 1 - (D^\bl-1)/(n(u) - 1)$.
}

\newcommand{\propositiontworv}{
$X$ and $Y$ are two random variables, and $f(X, Y)$ and $g(Y)$ are two functions. If $E_X f(X, Y) = 0$, then $\Var_{X, Y}\left[f(X, Y) + g(Y)\right]=\Var_{X, Y} f(X, Y) + \Var_Y g(Y)$.
}

\newcommand{\propositionvarr}{
If $\hat \nv^\bl(u)$ contains $D^\bl$ samples from the set $\nv(u)$ without  replacement, $x_1, \dots, x_V$ are random variables, $\forall v, \E\left[x_{v}\right]=0$ and $\forall v_1\ne v_2, \Cov\left[x_{v_1}, x_{v_2}\right]=0$, 
then
$\Var_{X, \hat \nv^\bl(u)}\left[\frac{n(u)}{D^\bl}\sum_{v\in \hat \nv^\bl(u)} x_v \right] 
= \frac{n(u)}{D^\bl} \sum_{v\in\nbr}\var{}{x_v}.$
}

\newcommand{\exacttesting}{
For a constant sequence of $W_i=W$ and any $i>LI$ (i.e., after $L$ epochs), the activations computed by CV are exact, i.e., $Z_{CV, i}^{(l)} = Z^{(l)}$ for each $l \in [L] $ and $H_{CV, i}^{(l)} = H^{(l)}$ for each $l \in [L-1]$. }

\newcommand{\convergencethm}{
Assume that (1) the activation $\sigma(\cdot)$ is $\rho$-Lipschitz, (2) the gradient of the cost function $\nabla_z f(y, z)$ is $\rho$-Lipschitz and bounded, (3) $\inorm{g_{CV,\Vc}(W)}$, $\inorm{g(W)}$, and $\inorm{\nabla \Lc(W)}$ are all bounded by $G>0$ for all $\hat P, \Vc$ and $W$. (4) The loss $\mathcal L(W)$ is $\rho$-smooth, i.e., $|\Lc(W_2)-\Lc(W_1)-\iprod{\nabla \Lc(W_1), W_2-W_1}|\le \frac{\rho}{2}\norm{W_2-W_1}_F^2\forall W_1, W_2$, where $\iprod{A, B}=\mbox{tr}(A^\top B)$ is the inner product of matrix $A$ and matrix $B$. 
Then, there exists  $K>0$, s.t., $\forall N>LI$, if 
we  run SGD for $R\le N$ iterations, where $R$ is chosen uniformly from $[N]_+$, we have
\begin{align*}
\E_{R} \norm{\nabla \Lc(W_R)}^2_F \le 2\frac{\Lc(W_1) - \Lc(W^*)+K+\rho K}{\sqrt N},
\end{align*}
for the updates $W_{i+1} = W_i - \gamma g_{CV,i}(W_i)$ and the step size $\gamma=\min\{\frac{1}{\rho},  \frac{1}{\sqrt{N}}\}$.
}
\section{Introduction}
\begin{figure*}
        \vspace{-1mm}
	\centering
	\includegraphics[width=\linewidth]{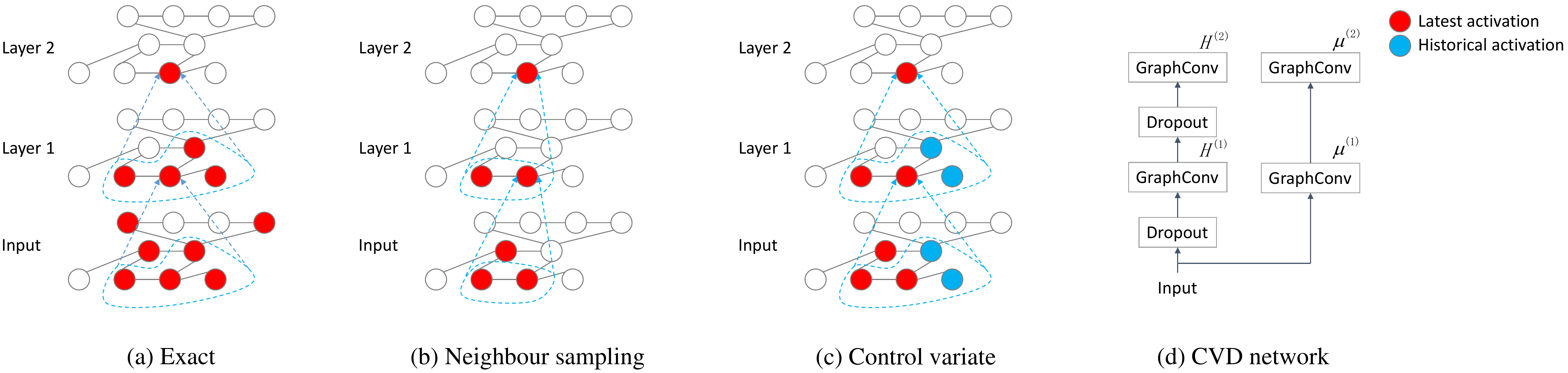}
	\vspace{-5mm}
	\caption{Two-layer graph convolutional networks, and the receptive field of a single vertex. }
	\label{fig:gcn-architecture}
\end{figure*} 

\setlength{\abovedisplayskip}{3pt}
\setlength{\abovedisplayshortskip}{1pt}
\setlength{\belowdisplayskip}{3pt}
\setlength{\belowdisplayshortskip}{1pt}
\setlength{\jot}{3pt}
\setlength{\textfloatsep}{3pt}  

Graph convolution networks (GCNs)~\citep{kipf2016semi} generalize convolutional neural networks (CNNs)~\citep{lecun1995convolutional} to graph structured data. The ``graph convolution'' operation applies same linear transformation to all the neighbors of a node, followed by mean pooling and nonlinearity. By stacking multiple graph convolution layers, GCNs can learn node representations by utilizing information from distant neighbors. GCNs and their variants~\cite{hamilton2017inductive,velivckovic2017graph} 
have been applied to semi-supervised node classification~\citep{kipf2016semi}, inductive node embedding~\citep{hamilton2017inductive}, link prediction~\citep{kipf2016variational,berg2017graph} and knowledge graphs~\citep{schlichtkrull2017modeling}, outperforming multi-layer perceptron (MLP) models that do not use the graph structure, and graph embedding approaches~\citep{perozzi2014deepwalk,tang2015line,grover2016node2vec} that do not use node features.

However, the graph convolution operation makes GCNs difficult to be trained efficiently. The representation of a node at layer $L$ is computed recursively by the representations of all its neighbors at layer $L-1$. Therefore, the receptive field of a single node grows exponentially with respect to the number of layers, as illustrated in Fig.~\ref{fig:gcn-architecture}(a). Due to the large receptive field size,~\citet{kipf2016semi} propose to train GCN by a batch algorithm, which computes the representations of all the nodes altogether. However, batch algorithms cannot handle large-scale datasets because of their slow convergence and the requirement to fit the entire dataset in GPU memory.

\citet{hamilton2017inductive} make an initial attempt to develop stochastic training algorithms for GCNs via a scheme of neighbor sampling (NS). Instead of considering all the neighbors, they randomly subsample $D^{(l)}$ neighbors at the $l$-th layer. Therefore, they reduce the receptive field size to $\prod_l D^{(l)}$, as shown in Fig.~\ref{fig:gcn-architecture}(b). They find that for two-layer GCNs, keeping $D^{(1)}=10$ and $D^{(2)}=25$ neighbors can achieve comparable performance with the original model.
However, there is no theoretical guarantee on the convergence of the stochastic training algorithm with NS. Moreover, the time complexity of NS is still $D^{(1)}D^{(2)}=250$ times larger than training an MLP, which is unsatisfactory.

In this paper, we develop novel control variate-based stochastic approximation algorithms for GCN. We utilize the historical activations of nodes as a control variate. We show that while the variance of the NS estimator depends on the magnitude of the activation, the variance of our algorithms only depends on the difference between the activation and its historical value. Furthermore, our algorithms bring new theoretical guarantees. At testing time, our algorithms give exact and zero-variance predictions, and at training time, our algorithms converge to a local optimum of GCN \emph{regardless of the neighbor sampling size $D^\bl$}.
The theoretical results allow us to significantly reduce the time complexity by sampling only two neighbors per node, yet still retain the quality of the model.

We empirically test our algorithms on six graph datasets, and show that our techniques significantly reduce the bias and variance of the gradient from NS with the same receptive field size. Despite sampling only $D^{(l)}=2$ neighbors, our algorithms achieve the same predictive performance with the exact algorithm in a comparable number of epochs on all the datasets, i.e., we reduce the time complexity while having almost no loss on the speed of convergence, which is the best we can expect. On the largest Reddit dataset, the training time of our algorithm is 7 times shorter than that of the best-performing competitor among the exact algorithm~\cite{kipf2016semi}, neighbor sampling~\cite{hamilton2017inductive} and importance sampling~\cite{chen2018fastgcn} algorithms.
\chris{Maybe we should emphize more about computing the average of neighbors, and mention more about our general applicability}

\begin{table}[t]
	\centering
	\small{
		\begin{tabular}{c|cccc}
\hline
Dataset  & $V$ & $E$ & Degree & Degree 2 \\
\hline
Citeseer &  3,327 & 12,431 & 4 & 15 \\
Cora     & 2,708 & 13,264 & 5 & 37 \\
PubMed   & 19,717 & 108,365 & 6 & 60 \\
NELL     & 65,755 & 318,135 & 5 & 1,597 \\
PPI      &  14,755 & 458,973 & 31 & 970 \\
Reddit   & 232,965 & 23,446,803 & 101 & 10,858 \\
\hline
		\end{tabular}
	}
	\caption{Number of vertexes, edges, and average number of 1-hop and 2-hop neighbors per node for each dataset. Undirected edges are counted twice and self-loops are counted once.}
	\label{tab:datasets}
\end{table}

\section{Backgrounds}

We now briefly review graph convolutional networks (GCNs), stochastic training, and the neighbor sampling (NS) and  importance sampling (IS) algorithms.

\subsection{Graph Convolutional Networks}\label{sec:gcn}
We present our algorithm with a GCN for semi-supervised node classification~\citep{kipf2016semi}. However, the algorithm is neither limited to the task nor the model. Our algorithm is applicable to other models~\cite{hamilton2017inductive} and tasks~\citep{kipf2016variational,berg2017graph,schlichtkrull2017modeling, hamilton2017representation} that involve computing the average activation of neighbors.

In the node classification task, we have an undirected graph $\mathcal G=(\mathcal V, \mathcal E)$ with $V=|\mathcal V|$ vertices and $E=|\mathcal E|$ edges, where each vertex $v$ consists of a feature vector $x_v$ and a label $y_v$. We observe the labels for some vertices $\Vc_\Lc$. The goal is to predict the labels for the rest vertices $\Vc_\Uc:=\mathcal V\backslash \Vc_\Lc$. The edges are represented as a symmetric $V\times V$ adjacency matrix $A$, where $A_{uv}$ is the weight of the edge between $u$ and $v$, and the propagation matrix $P$ is a normalized version of $A$:
$\tilde A = A + I$, $\tilde D_{uv}=\sum_{v} \tilde A_{uv}$, and $P = \tilde D^{-\frac{1}{2}} \tilde A \tilde D^{-\frac{1}{2}}$.
A graph convolution layer is defined as 
\begin{align}\label{eqn:gcn}
Z^{(l+1)}=P  H^{(l)} W^{(l)}, \quad H^{(l+1)}=\sigma(Z^{l+1}),
\end{align}
where $H^{(l)}$ is the activation matrix in the $l$-th layer, whose each row is the activation of a graph node. $H^{(0)}=X$ is the input feature matrix, $W^{(l)}$ is a trainable weight matrix, and $\sigma(\cdot)$ is an activation function. Denote $\abs{\cdot}$ as the cardinality of a set. The training loss is defined as
\begin{align}
\Lc=\frac{1}{|\Vc_\Lc|}\sum_{v\in \Vc_\Lc} f(y_v, z^{(L)}_v),\label{eqn:loss}
\end{align}
where $f(\cdot, \cdot)$ is a loss function.
A graph convolution layer propagates information to nodes from their neighbors by computing the neighbor averaging $PH^{(l)}$. Let $\nv(u)$ be the set of neighbors of node $u$, and $n(u)$ be its cardinality, the neighbor averaging of node $u$, $(PH^{(l)})_u=\sum_{v=1}^V P_{uv} h^{(l)}_v=\sum_{v\in \nv(u)} P_{uv} h^{(l)}_v$, is a weighted sum of neighbors' activations. Then, a fully-connected layer is applied on all the nodes, with a shared weight matrix $W^{(l)}$ across all the nodes.

We denote the \emph{receptive field} of node $u$ at layer $l$ as all the activations $h_v^\bl$ on layer $l$ needed for computing $z_u^\bL$. If the layer is not explicitly mentioned, it means layer 0. The receptive field of node $u$ is all its $L$-hop neighbors, i.e., nodes that are reachable from $u$ within $L$ hops, as illustrated in Fig.~\ref{fig:gcn-architecture}(a). 
When $P=I$, GCN reduces to a multi-layer perceptron (MLP) model which does not use the graph structure. For MLP, the receptive field of a node $u$ is just the node itself.

\subsection{Stochastic Training}
It is generally expensive to compute the batch gradient $\nabla \Lc=\frac{1}{|\Vc_\Lc|}\sum_{v\in \Vc_\Lc} \nabla f(y_v, z^{(L)}_v)$, which involves iterating over the entire labeled set of nodes. A possible solution is to approximate the batch gradient by a stochastic gradient 
\begin{align}
  \frac{1}{|\Vc_\Bc|}\sum_{v\in \Vc_\Bc} \nabla f(y_v, z^{(L)}_v), \label{eqn:sgd}   
\end{align}
where $\Vc_\Bc \subset \Vc_\Lc$ is a minibatch of labeled nodes. However, this gradient is still expensive to compute, due to the large receptive field size. For instance, as shown in Table~\ref{tab:datasets}, the number of 2-hop neighbors on the NELL dataset is averagely 1,597, which means computing the gradient for a single node in a 2-layer GCN involves touching $1,597/65,755\approx 2.4\%$ nodes of the entire graph. 

In subsequent sections, two other stochasticity will be introduced besides the random selection of the minibatch: the random sampling of neighbors (Sec.~\ref{sec:neighbor-subsampling}) and the random dropout of features (Sec.~\ref{sec:dropout}).

\subsection{Neighbor Sampling}\label{sec:neighbor-subsampling}

To reduce the receptive field size,~\citet{hamilton2017inductive} propose a neighbor sampling (NS) algorithm. NS randomly chooses $D^{(l)}$ neighbors for each node at layer $l$ and develops an estimator $\NS_u^\bl$ of $(PH^{(l)})_u$ based on Monte-Carlo approximation:
$$(PH^{(l)})_u\approx \NS_u^\bl:=\frac{n(u)}{D^{(l)}} \sum_{v\in \hat\nv^\bl(u)} P_{uv} h_v^\bl,$$ where
$\hat\nv^\bl(u)\subset \nv(u)$ is a subset of  $D^{(l)}$ random neighbors.
Therefore, NS reduces the receptive field size from all the $L$-hop neighbors to the number of sampled neighbors, $\prod_{l=1}^L D^{(l)}$. 
We refer $\NS_u^\bl$ as the NS estimator of $(PH^\bl)_u$, and $(PH^\bl)_u$ itself as the exact estimator.

Neighbor sampling can also be written in a matrix form as
\begin{align}\label{eqn:mcgcn}
Z^{(l+1)}=\hat P^{(l)} H^{(l)} W^{(l)}, 
\quad H^{(l+1)}=\sigma(Z^{(l+1)}),
\end{align}
where the propagation matrix $P$ is replaced by a sparser unbiased estimator $\hat P^{(l)}$, i.e., $\E \hat P^{(l)} = P$,
where $\hat P^{(l)}_{uv}=\frac{n(u)}{D^\bl}P_{uv}$ if $v\in \hat\nv^\bl(u)$, and $\hat P^{(l)}_{uv}=0$ otherwise. \citet{hamilton2017inductive} propose to perform the approximate forward propagation as Eq.~(\ref{eqn:mcgcn}), and do stochastic gradient descent (SGD) with the auto-differentiation gradient. The approximated gradient has two sources of randomness: the random selection of minibatch $\mathcal V_\Bc\subset \Vc_\Lc$, and the random selection of neighbors.

Though $\hat P^{(l)}$ is an unbiased estimator of $P$, $\sigma(\hat P^{(l)} H^{(l)} W^{(l)})$ is \emph{not} an unbiased estimator of $\sigma(P H^{(l)} W^{(l)})$, due to the non-linearity of $\sigma(\cdot)$. In the sequel, both the prediction $Z^\bL$ and gradient $\nabla f(y_v, z_v^\bL)$ obtained by NS are biased, and the convergence of SGD is not guaranteed, unless the sample size $D^\bl$ goes to infinity. Because of the biased gradient, the sample size $D^\bl$ needs to be large for NS, to keep comparable predictive performance with the exact algorithm. \citet{hamilton2017inductive} choose  $D^{(1)}=10$ and $D^{(2)}=25$, and the receptive field size $D^{(1)}\times D^{(2)}=250$ is much larger than that of MLP, which is $1$, so the training is still expensive.

\subsection{Importance Sampling}\label{sec:importance-sampling}
FastGCN~\citep{chen2018fastgcn} is another sampling-based algorithm similar as NS. Instead of sampling neighbors for each node, FastGCN directly subsample the receptive field for each layer altogether. Formally, it approximates $(PH^\bl)_u$ with $S$ samples $v_1, \dots, v_S\in \Vc$ as
$$(PH^\bl)_u=V\sum_{v=1}^V\frac{1}{V} P_{uv}h_v^\bl\approx\frac{V}{S}\sum_{v_s\sim q(v)} P_{uv}h_{v_s}^\bl/q(v_s),$$
where the importance distribution $q(v)\propto \sum_{u=1}^V P_{uv}^2=\frac{1}{n(v)}\sum_{(u,v)\in\Ec} \frac{1}{n(u)}$, according the definition of $P$ in Sec.~\ref{sec:gcn}. We refer to this estimator as importance sampling (IS).~\citet{chen2018fastgcn} show that IS performs better than using a uniform sample distribution $q(v)\propto 1$. NS can be viewed as an IS estimator with the importance distribution $q(v)\propto \sum_{(u, v)\in \Ec} \frac{1}{n(u)}$, because each node $u$ has probability $\frac{1}{n(u)}$ to choose the neighbor $v$. Though IS may have a smaller variance than NS, it still only guarantees the convergence as the sample size $S$ goes to infinity. Empirically, we find IS to work even \emph{worse} than NS because sometimes it can select many neighbors for one node, and no neighbor for another, in which case the activation of the latter node is just meaningless zero.

\section{Control Variate Based Algorithm}

We  present a novel control variate based algorithm that utilizes  historical activations to reduce the estimator variance.
\junz{algorithms or an algorithm? keep consistent}

\subsection{Control Variate Based Estimator}\label{sec:cv}
While computing the neighbor average $\sum_{v\in \nv(u)} P_{uv} h^{(l)}_v$, we cannot afford to evaluate all the $h^{(l)}_v$ terms because they need to be computed recursively, i.e., we again need the activations $h^{(l-1)}_w$ of all of $v$'s neighbors $w$. 

Our idea is to maintain the history $\bar h_v^\bl$ for each  $h_v^\bl$ as an affordable approximation.
Each time when $h_v^\bl$ is computed, we update $\bar h_v^\bl$ with $h_v^\bl$. We expect $\bar h_v^\bl$ and $h_v^\bl$ to be similar if the model weights do not change too fast during the training.
Formally, let $\Delta h_v^\bl=h_v^\bl - \bar h_v^\bl$, we approximate
\begin{align}
&(PH^{(l)})_u = \sum_{v\in \nv(u)} P_{uv} \Delta h^{(l)}_v + \sum_{v\in \nv(u)} P_{uv} \bar h^{(l)}_v\approx \CV_u^\bl \nonumber \\
&:=\frac{n(u)}{D^{(l)}} \sum_{v\in \hat\nv^\bl (u)} P_{uv} \Delta h^{(l)}_v + \sum_{v\in \nv(u)} P_{uv} \bar h^{(l)}_v,\label{eqn:cv}
\end{align}
where we represent $h_v^\bl$ as the sum of $\Delta h_v^\bl$ and $\bar h_v^\bl$, and we only apply Monte-Carlo approximation on the $\Delta h_v^\bl$ term. Averaging over all the $\bar h_v^\bl$'s is still affordable because they do not need to be computed recursively.
Since we expect $h_v^\bl$ and $\bar h_v^\bl$ to be close, $\Delta h_v$ will be small and $\CV_u^\bl$ should have a smaller variance than $\NS_u^\bl$. Particularly, if the model weight is kept fixed, $\bar h_v^\bl$ should  eventually equal with $h_v^\bl$, so that $\CV_u^\bl= 0 + \sum_{v\in \nv(u)} P_{uv} \bar h^{(l)}_v=\sum_{v\in \nv(u)} P_{uv} h^{(l)}_v = (PH^{(l)})_u$, i.e., the estimator has zero variance. 
This estimator is referred as CV.
We will compare the variance of NS and CV estimators in Sec.~\ref{sec:var1} and show that the variance of CV will be eventually zero during the training in Sec.~\ref{sec:theory}.
The term $\CV_u^\bl-\NS_u^\bl=\sum_{v\in \nv(u)} P_{uv}\bar h_u^\bl - \frac{n(u)}{D^{(l)}} \sum_{v\in \hat\nv^\bl (u)} P_{uv} \bar h_u^\bl$  is a \emph{control variate}~\citep[Chapter~5]{ripley2009stochastic} added to the neighbor sampling estimator $\NS_u^\bl$, to reduce its variance. 

In matrix form, let $\bar H^\bl$ be the matrix formed by stacking $\bar h_v^\bl$, then CV can be written as
\begin{align}
Z^{(l+1)}=&\left(\hat P^{(l)} (H^\bl-\bar H^\bl) + P \bar H^\bl\right) W^{(l)}.
\label{eqn:zupdate}
\end{align}

\subsection{Variance Analysis}\label{sec:var1}

We analyze the variance of the estimators assuming all the features are 1-dimensional. The analysis can be extended to multiple dimensions by treating each dimension separately. 
We further assume that $\snbr$ is created by sampling $D^\bl$ neighbors without replacement from $\nbr$. The following proposition is proven in Appendix~\ref{S-sec:variance-derivation}:
\begin{proposition}\label{proposition:var}
\propositionvar
\end{proposition}

By Proposition~\ref{proposition:var}, we have $\Var_{\hat \nv^\bl(u)}\left[ NS_u^\bl \right] 
=\frac{C_u^\bl}{2D^\bl}  \sum_{v_1\in \nv(u)}\sum_{v_2\in\nv(u)} (P_{uv_1}h_{v_1}^\bl-P_{uv_2}h_{v_2}^\bl)^2
$, which is the total distance of the weighted activations of all pairs of neighbors, and is zero iff $P_{uv}h_{v}$ is identical for all neighbors, in which case any neighbor contains all the information of the entire neighborhood. 

The variance of the CV estimator is $\Var_{\hat \nv^\bl(u)} \left[CV_u^\bl\right] = 
\frac{C_u^\bl}{2D^\bl}  \sum_{v_1\in \nv(u)}\sum_{v_2\in\nv(u)} (P_{uv_1}\Delta h_{v_1}^\bl-P_{uv_2}\Delta h_{v_2}^\bl)^2$, which replaces $h_v^\bl$ by $\Delta h_v^\bl$. 
Since $\Delta h_v^\bl$ is usually much smaller than $h_v^\bl$, the CV estimator enjoys much smaller variance than the NS estimator. Furthermore, as we will show in Sec.~\ref{sec:convergence}, $\Delta h_v^\bl$ converges to zero during training, so we achieve not only \emph{variance reduction} but \emph{variance elimination}, as  the variance vanishes eventually.

\subsection{Implementation Details and Time Complexity}\label{sec:implementation}

Training with the CV estimator is similar as with the NS estimator~\cite{hamilton2017inductive}. Particularly, each iteration of the algorithm involves the following steps: 
\fbox{
 \parbox{0.95\linewidth}{
{\bf Stochastic GCN with Variance Reduction}  
\begin{enumerate}[noitemsep,nolistsep,leftmargin=*,topsep=0pt]
  \item Randomly select a minibatch $\mathcal V_B\in \mathcal V_L$ of nodes; 
  \item Build a computation graph that only contains the activations $h_v^\bl$ and $\bar h_v^\bl$ needed for the current minibatch; 
  \item Get the predictions by forward propagation as Eq.~(\ref{eqn:zupdate});
  \item Get the gradients by backward propagation, and update the parameters by SGD;
  \item Update the historical activations.
\end{enumerate}
}
}
%
%
Step 3 and 4 are handled automatically by frameworks such  de TensorFlow~\cite{abadi2016tensorflow}. The computational graph at Step 2 is defined by the receptive field $\rv^\bl$ and the propagation matrices $\hat P^\bl$ at each layer. The receptive field $\rv^\bl$ specifies the  activations $h_v^\bl$ of which nodes should be computed for the current minibatch, according to Eq.~(\ref{eqn:zupdate}).
We can construct $\rv^\bl$ and $\hat P^\bl$ from top to bottom, by randomly adding $D^\bl$ neighbors for each node in $\rv^\bll$, starting with $\rv^\bL=\mathcal V_\Bc$. We assume $h_v^\bl$ is always needed to compute $h_v^\bll$, i.e., $v$ is always selected as a neighbor of itself. The receptive fields are  illustrated in Fig.~\ref{fig:gcn-architecture}(c), where red nodes are in receptive fields, whose activations $h_v^\bl$ are needed, and the histories $\bar h_v^\bl$ of blue nodes are also needed. Finally, in Step 5, we update $\bar h_v^\bl$ with $h_v^\bl$ for each $v\in \rv^\bl$. We have the pseudocode for the training in Appendix~\ref{S-sec:pseudocode}.

GCN has two main types of computation, namely, the sparse-dense matrix multiplication (SPMM) such as $PH^\bl$, and the dense-dense matrix multiplication (GEMM) such as $U W^\bl$. We assume that the node feature is $K$-dimensional and the first hidden layer is $A$-dimensional. 

For batch GCN, the time complexity is $O(EK)$ for SPMM and $O(VKA)$ for GEMM. 
For our stochastic training algorithm with control variates, 
the dominant SPMM computation is the average of neighbor history $P\bar H^{(0)}$ for the nodes in $\rv^{(1)}$, whose size is $O(|\mathcal V_B|\prod_{l=2}^L D^\bl)$. For example, in a 2-layer GCN where we sample $D^\bl=2$ neighbors for each node, $\prod_{l=2}^L D^\bl=2$.
Therefore, the time complexity of SPMM is $O(V D K \prod_{l=2}^L D^\bl)$ per epoch, where $D$ is the average degree of nodes in $\rv^{(1)}$.\footnote{$VD\ne E$ because the probability of each node to present in $\rv^{(1)}$ is different. Nodes of higher degree have larger probability to present. We can also subsample neighbors' history if $D$ is large.}
The dominant GEMM computation is the first fully-connected layer on all the nodes in $\rv^{(1)}$, whose time complexity is $O(VKA \prod_{l=2}^L D^\bl)$ per epoch.


\section{Theoretical Results}\label{sec:theory}

Besides smaller variance, CV also has stronger theoretical guarantees than NS. In this section, we present two theorems. One states that if the model parameters are fixed, e.g., during testing, CV produces exact predictions after $L$ epochs; and the other establishes the convergence towards a local optimum regardless of the neighbor sampling size.

In this section, we assume that the algorithm is run by epochs. In each epoch, we randomly partition the vertex set $\mathcal V$ as $I$ minibatches $\mathcal V_1, \dots, \mathcal V_{I}$, and in the $i$-th iteration, we run a forward pass to compute the predictions for nodes in $\mathcal V_i$, an optional back propagation to compute the gradients, and update the history. 
Note that in each epoch we scan \emph{all} the nodes instead of just training nodes, to ensure that the history of each node is updated at least once per epoch.

We denote the model parameters in the $i$-th iteration as $W_i$. At training time,  $W_i$ is updated by SGD over time; at testing time, $W_i$ is kept fixed. To distinguish, the activations produced by CV at iteration $i$ are denoted as $Z_{CV, i}^\bl$ and $H_{CV, i}^\bl$, and the activations produced by the exact algorithm (Eq.~\ref{eqn:gcn}) are denoted as $Z_{i}^\bl$ and $H_{i}^\bl$. At iteration $i$, the network computes the predictions and gradients for the minibatch $\Vc_i$, where $g_{CV,i}(W_i):=\frac{1}{|\Vc_i|}\sum_{v\in \Vc_i} \nabla f(y_v, z_{CV,i,v}^\bL)$ and $g_{i}(W_i):=\frac{1}{|\Vc_i|}\sum_{v\in \Vc_i} \nabla f(y_v, z_{i,v}^\bL)$ are the stochastic gradients computed by CV and the exact algorithm. $\nabla \Lc(W_i)=\frac{1}{|\Vc_{\Lc}|}\sum_{v\in \Vc_{\Lc}} \nabla f(y_v, z_{v}^\bL)$ is the deterministic batch gradient computed by the exact algorithm. The subscript $_i$ may be omitted for the exact algorithm if $W_i$ is a constant sequence. We let $[L]=\{0, \dots, L\}$ and $[L]_+=\{1, \dots, L\}$.
The gradient $g_{CV, i}(W_i)$ has two sources of randomness: the random selection of the minibatch $\mathcal V_i$ and randomness of the neighbors $\hat P$, so we may take expectation of  $g_{CV, i}(W_i)$ w.r.t. either $\mathcal V_i$ or $\hat P$, or both.

\subsection{Exact Testing}\label{sec:prediction}
The following theorem reveals the connection of 
the exact and approximate predictions by CV.
\begin{theorem}\label{thm:prediction}
\exacttesting
\end{theorem}
Theorem~\ref{thm:prediction} shows that at testing time, we can run forward propagation with CV for $L$ epoches and get exact prediction.
This outperforms NS, which cannot recover the exact prediction unless the neighbor sample size goes to infinity. Comparing with directly making exact predictions by an exact batch algorithm, CV is more scalable because it does not need to load the entire graph into memory. The proof can be found in Appendix~\ref{S-sec:proof-of-thm1}.

\subsection{Convergence Guarantee}\label{sec:convergence}

The following theorem shows that SGD training with the approximated gradients $g_{CV, i}(W_i)$ still  converges to a local optimum, regardless of the neighbor sampling size $D^{(l)}$. Therefore, we can choose arbitrarily small $D^{(l)}$ without worrying about the convergence.
\begin{theorem}\label{thm:convergence}
\convergencethm
\end{theorem}
Particularly, $\lim_{N\rightarrow \infty} \E_{R}  \norm{\nabla \mathcal L(W_R)}^2 = 0$. Therefore, our algorithm converges to a local optimum as the max number of iterations $N$ goes to infinity.
The full proof is in Appendix~\ref{S-sec:proof-of-thm2}. For short, we show that $g_{CV, i}(W_i)$ is unbiased as $i\rightarrow \infty$, and then show that SGD with such asymptotically unbiased gradients   converges to a local optimum.


\section{Handling Dropout of Features}\label{sec:dropout}
In this section, we consider introducing a third source of randomness,
the random dropout of features~\citep{srivastava2014dropout}. Let $\mbox{Dropout}_p(X) = M \circ X$ be the dropout operation, where $M_{ij}\sim \mbox{Bern}(p)$ are i.i.d. Bernoulli random variables, and $\circ$ is the element-wise product. Let $\E_M$ be the expectation over dropout masks.

With dropout, all the activations $h_v^\bl$ are random variables whose randomness comes from dropout, even in the exact algorithm Eq.~(\ref{eqn:gcn}). We want to design a cheap estimator for the random variable $(PH^\bl)_u=\sum_{v\in \nv(u)} P_{uv}h_v^\bl$, based on a stochastic neighborhood $\hat \nv^\bl(u)$.
An ideal estimator should have the same distribution with $(PH^\bl)_u$. 
However, such an estimator is difficult to design. Instead, we develop an estimator $\CVD_u^\bl$ that eventually has the same mean and variance with $(PH^\bl)_u$, i.e., $\E_{\hat \nv^\bl(u)} \E_M \CVD_u^\bl = \E_M (PH^\bl)_u$ and $\Var_{\hat \nv^\bl(u)} \Var_M \CVD_u^\bl = \Var_M (PH^\bl)_u$.

\subsection{Control Variate for Dropout}\label{sec:cvd}
With dropout, $\Delta h_v^\bl = h_v^\bl - \bar h_v^\bl$ is not necessarily small even if $\bar h_v^\bl$ and $h_v^\bl$ have the same distribution. We develop another stochastic approximation algorithm, \emph{control variate for dropout} (CVD), that works well with dropout.

Our method is based on the weight scaling procedure~\citep{srivastava2014dropout} to approximately compute the mean $\mu_v^\bl:=\E_M \left[ h_v^\bl \right]$. That is, along with the dropout model, we can run a copy of the model without dropout to obtain the mean $\mu_v^\bl$, as illustrated in Fig.~\ref{fig:gcn-architecture}(d). We obtain a stochastic approximation by separating the mean and variance

\begin{align*}
&(PH^{(l)})_u = \sum_{v\in \nv(u)} P_{uv} (\rh^\bl_v + \Delta \mu^\bl_v + \bar\mu^\bl_v) \approx \CVD_u^\bl  \nonumber \\
&:=\sqrt{R} \sum_{v\in\hat\nv} P_{uv} \rh^\bl_v  + 
R   \sum_{v\in \hat\nv} P_{uv} \Delta \mu^\bl_v 
+ \sum_{v\in \nv(u)} P_{uv}\bar\mu^\bl_v,
\end{align*}
where $\nv=\snbr$, $R=n(u)/D^{(l)}$ for short, $\rh_v^\bl=h_v^\bl-\mu_v^\bl$, $\bar\mu_v^\bl$ is the historical mean activation, obtained by storing $\mu_v^\bl$ instead of $h_v^\bl$, and $\Delta\mu_v^\bl = \mu_v^\bl - \bar\mu_v^\bl$. We separate $h_v^\bl$ as three terms, the latter two terms on $\mu_v^\bl$ do not have the randomness from dropout, and $\mu_v^\bl$ are treated as if $h_v^\bl$ for the CV estimator.  The first term has zero mean w.r.t. dropout, i.e., $\E_M \rh_v^\bl=0$. 
We have $\E_{\hat \nv^\bl(u)} \E_M \CVD_u^\bl = 0 + \sum_{v\in \nv(u)} P_{uv}(\Delta\mu_v^\bl+\bar\mu_v^\bl) = \E_M (PH^\bl)_u$, i.e., the estimator is unbiased, and we shall see that the estimator eventually has the correct variance if $h_v^\bl$'s are uncorrelated in Sec.~\ref{sec:variance-analysis}.

\subsection{Variance Analysis}\label{sec:variance-analysis}
\begin{table}[t]
	\centering
	\begin{tabular}{ccc}
		\toprule
		Esti.            & VNS & VD \\   
		\midrule
		Exact      & 0                       & $S_u^\bl$            \\   
		NS         &   $ (P_{uv_1} \mu_{v_1}^\bl-P_{uv_2}\mu_{v_2}^\bl)^2$ 
		& $\frac{n(u)}{D^\bl}S_u^\bl $ \\  
		CV         & $ (P_{uv_1}\Delta \mu_{v_1}^\bl-P_{uv_2}\Delta \mu_{v_2}^\bl)^2$  &
		$\left(3+\frac{n(u)}{D^\bl}\right)S_u^\bl $ \\   
		CVD        &
		$ (P_{uv_1}\Delta \mu_{v_1}^\bl-P_{uv_2}\Delta \mu_{v_2}^\bl)^2$ & $S_u^\bl$
		\\
		\bottomrule
	\end{tabular}
	\vspace{-3mm}
	\caption{Variance of different estimators. To save space we omit  $\frac{C_u^\bl}{2D^\bl}  \sum_{v_1, v_2\in \nv(u)}$ before all the VNS terms.}
	\label{tab:gaussian-variance}
\end{table}

We analyze the variance under the assumption that the node activations are uncorrelated, i.e., $\Cov_M\left[ h_{v_1}^\bl, h_{v_2}^\bl \right]=0, \forall v_1\ne v_2$. We report the correlation between nodes empirically in Appendix~\ref{S-sec:indep-gaussian}.
To facilitate the analysis of the variance, we introduce two propositions proven in Appendix~\ref{S-sec:variance-derivation} . The first  helps the derivation of the dropout variance; and the second implies that we can treat the variance introduced by neighbor sampling and by dropout separately. 
\begin{proposition}\label{proposition-varr}
	\propositionvarr
\end{proposition}
\begin{proposition}\label{proposition-varsum}
\propositiontworv
\end{proposition}

By Proposition~\ref{proposition-varsum}, $\Var_{\hat \nv} \Var_M \CVD_u^\bl$ can be written as the sum of 
$\Var_{\hat \nv} \Var_M 
\left[\sqrt{R} \sum_{v\in \hat\nv} P_{uv} \rh^\bl_v\right]$ and $\Var_{\hat \nv} \left[ R   \sum_{v\in \hat\nv} P_{uv} \Delta \mu^\bl_v
+\sum_{v\in \nv(u)} P_{uv}\bar\mu^\bl_v \right].$ We refer the first term as the variance from dropout (VD) and the second term as the variance from neighbor sampling (VNS). Ideally, VD should equal to the variance of $(PH^\bl)_u$ and VNS should be zero.
VNS can be derived by replicating the analysis in Sec.~\ref{sec:var1}, replacing $h$ with $\mu$. 
Let $s_v^\bl=\Var_M h_v^\bl=\Var_M \rh_v^\bl$, and $S^\bl_u=\Var_M (PH^\bl)_u=
\sum_{v\in \nv(u)} P_{uv}^2 s_v^\bl$, 
By Proposition~\ref{proposition-varr}, VD of $\CVD_u^\bl$, 
$\sum_{v\in\nbr}P_{uv}^2 \var{}{\rh^\bl_v}=S_u^\bl$, equals with the VD of the exact estimator as desired.

We summarize the estimators and their variances in Table~\ref{tab:gaussian-variance}, where the derivations are in Appendix~\ref{S-sec:variance-derivation}. As in Sec.~\ref{sec:var1}, VNS of CV and CVD depends on $\Delta \mu_v$, which converges to zero as the training progresses, while VNS of NS depends on the non-zero $\mu_v$.
On the other hand, CVD is the only estimator except the exact one that gives correct VD.

\subsection{Preprocessing Strategy}\label{sec:preprocessing}
There are two possible models adopting dropout, $Z^{(l+1)}=P \mbox{Dropout}_p(H^{(l)}) W^{(l)}$ or $Z^{(l+1)}=\mbox{Dropout}_p(P H^{(l)}) W^{(l)}$. The difference is whether the dropout layer is before or after neighbor averaging. \citet{kipf2016semi} adopt the former one, and we adopt the latter one, while the two models performs similarly in practice, as we shall see in Sec.~\ref{sec:experiment-preprocessing}. The advantage of the latter model is that we can preprocess $U^{(0)}=PH^{(0)}=PX$ and takes $U^{(0)}$ as the new input. In this way, the actual number of graph convolution layers is reduced by one --- the first layer is merely a fully-connected layer instead of a graph convolution one. Since most GCNs only have two graph convolution layers~\citep{kipf2016semi,hamilton2017inductive}, this gives a significant reduction of the receptive field size and speeds up the computation. We refer this optimization as the preprocessing strategy. 
\section{Experiments}

We examine the variance and convergence of our algorithms empirically on six datasets, including Citeseer, Cora, PubMed and NELL from~\citet{kipf2016semi} and Reddit, PPI from~\citet{hamilton2017inductive}, as summarized in  Table~\ref{tab:datasets}, with the same train / validation / test splits.
To measure the predictive performance, we report Micro-F1 for the multi-label PPI dataset, and accuracy for all the other multi-class datasets.
The model is GCN for the former 4 datasets and GraphSAGE~\citep{hamilton2017inductive} for the latter 2 datasets, see Appendix~\ref{S-sec:experimental-setup} for the details on the architectures. We repeat the convergence experiments 10 times on Citeseer, Cora, PubMed and NELL, and 5 times on Reddit and PPI.  The experiments are done on a Titan X (Maxwell) GPU.

\begin{table}[t]
	\centering
		\begin{tabular}{c|ccc}
			\hline
Dataset     & M0 & M1 &  M1+PP \\
\hline
Citeseer 	& $70.8\pm .1$ & $70.9\pm .2$ & $70.9\pm .2$ \\
Cora     	& $81.7\pm .5$ & $82.0\pm .8$ & $81.9\pm .7$\\
PubMed	& $79.0\pm .4$ & $78.7\pm .3$ & $78.9\pm .5$\\
NELL		& - & $64.9\pm 1.7$ & $64.2\pm 4.6$  \\
PPI			&$97.9\pm .04$ & $97.8\pm .05$ & $97.6\pm .09$ \\
Reddit		&$96.2\pm .04$ & $96.3\pm .07$ & $96.3\pm .04$\\			
			\hline
		\end{tabular}
	\vspace{-3mm}
	\caption{Testing accuracy of different algorithms and models after fixed number of epochs. Our implementation does not support M0 on NELL so the result is not reported.}\vspace{-.2cm}
	\label{tab:quality}
\end{table}

\begin{figure}[t]
	\centering
	\includegraphics[width=\linewidth]{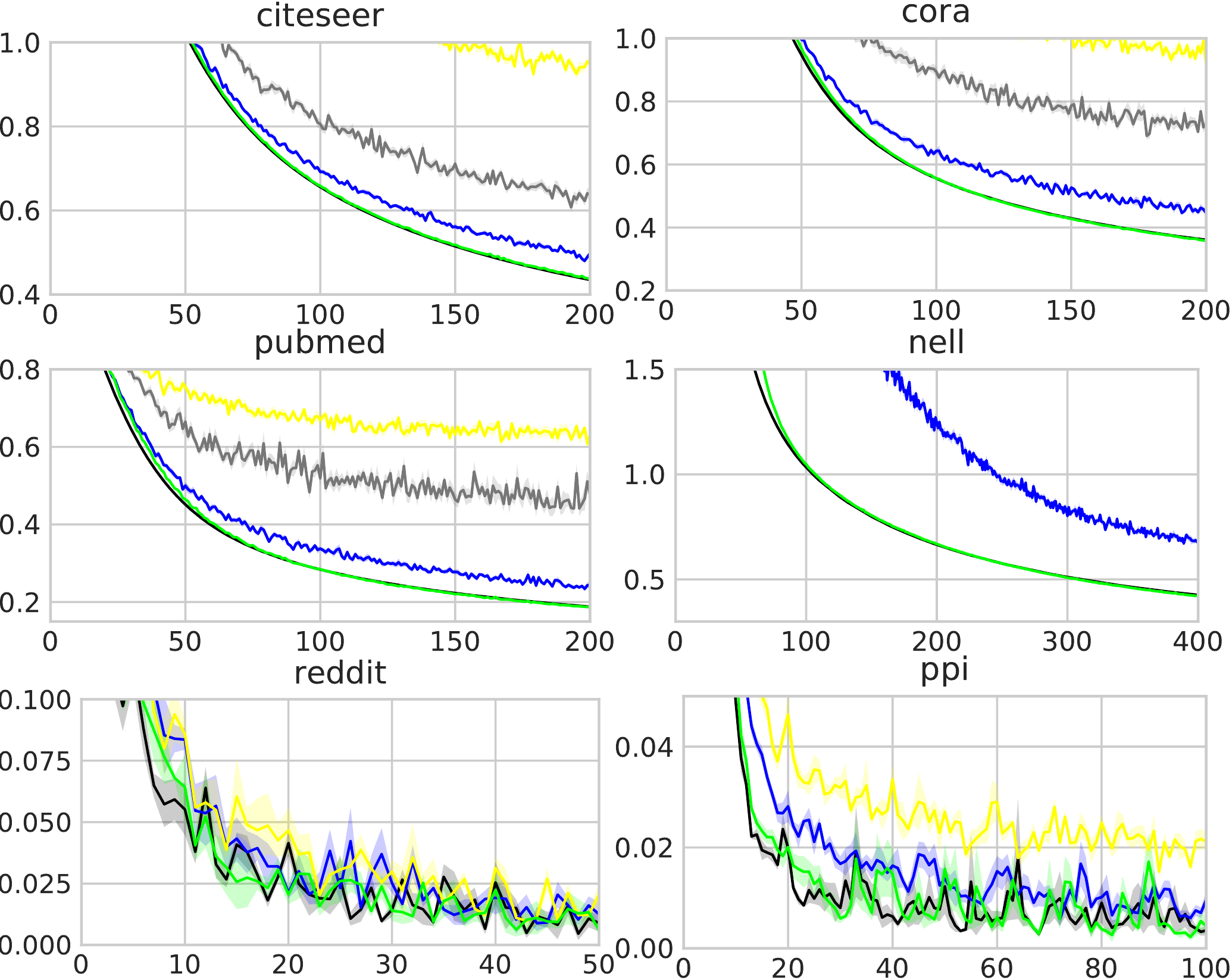}
	\includegraphics[width=\linewidth]{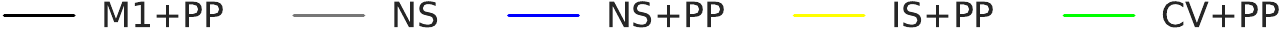}
        \vspace{-6mm}	
	\caption{Comparison of training loss with respect to number of epochs without dropout. The CV+PP curve overlaps with the Exact curve in the first four datasets. The training loss of NS and IS+PP are not shown on some datasets because they are too high.}
	\label{fig:iter-trainloss}
\end{figure}
\begin{figure}
	\centering
	\includegraphics[width=\linewidth]{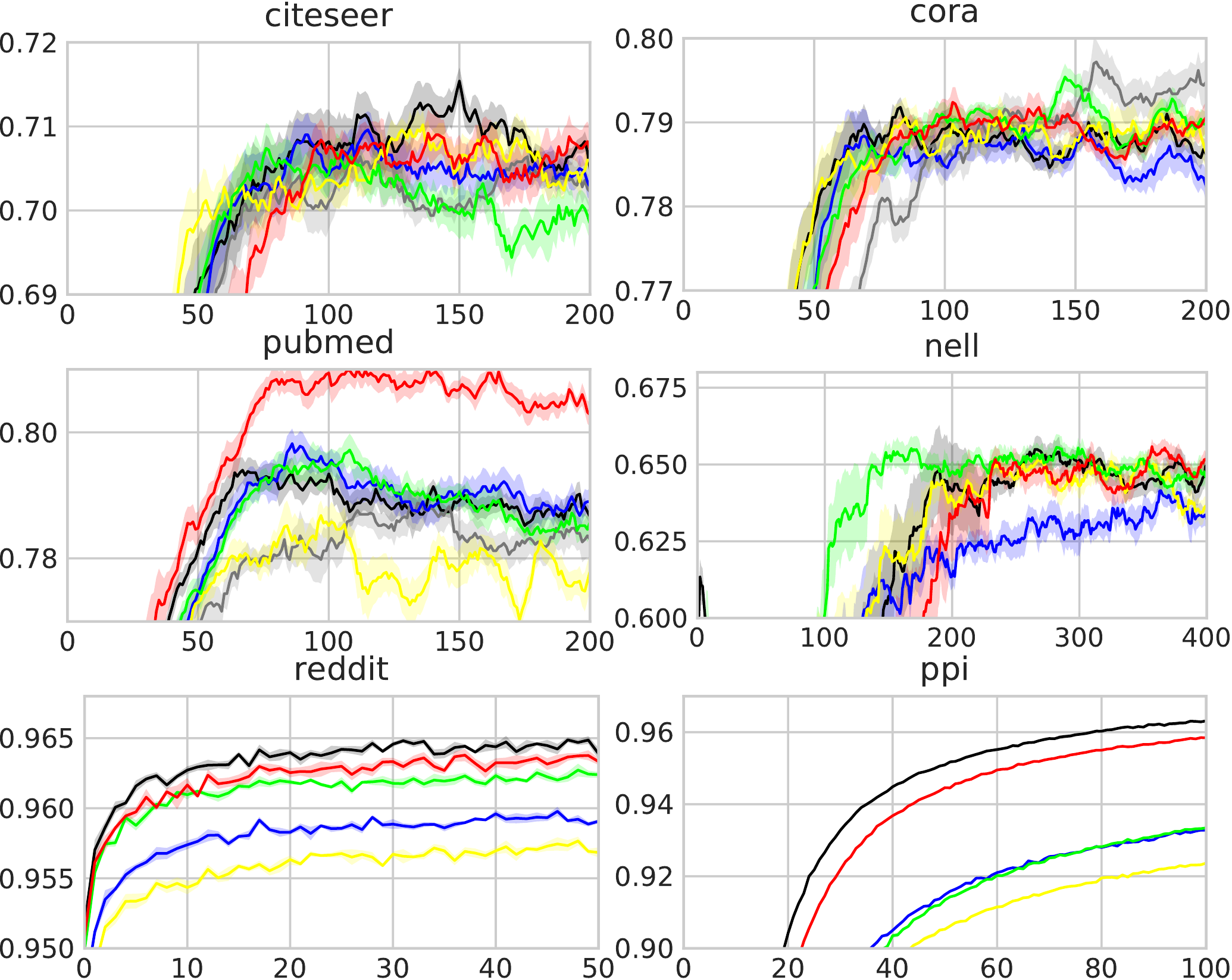}
	\includegraphics[width=\linewidth]{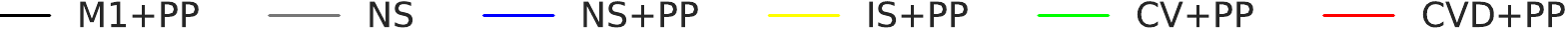}
        \vspace{-6mm}	
	\caption{Comparison of validation accuracy with respect to number of epochs. NS converges to 0.94 on the Reddit dataset and 0.6 on the PPI dataset.}
	\label{fig:iter-acc}
\end{figure}

\subsection{Impact of Preprocessing}\label{sec:experiment-preprocessing}
We first examine the impact of switching the order of dropout and computing neighbor averaging in Sec.~\ref{sec:preprocessing}. Let M0 be the $Z^{(l+1)}=P \mbox{Dropout}_p(H^{(l)}) W^{(l)}$ model by~\cite{kipf2016semi}, and M1 be our 
$Z^{(l+1)}= \mbox{Dropout}_p(P H^{(l)}) W^{(l)}$ model, we compare three settings: M0 and M1 are exact algorithms without any neighbor sampling, and M1+PP samples a large number of $D^{(l)}=20$ neighbors and preprocesses $PH^{(0)}$ so that the first neighbor averaging is exact.
In Table~\ref{tab:quality} we can see that all the three settings performs similarly, i.e., switching the order does not affect the predictive performance. Therefore, we use the fastest M1+PP  as the exact baseline in following convergence experiments.

\subsection{Convergence Results}\label{sec:convergence-dropout}
Having the M1+PP algorithm as an exact baseline, the next goal is reducing the time complexity per epoch to make it comparable with the time complexity of MLP, by setting $D^\bl=2$.
We cannot set $D^\bl=1$ because GraphSAGE explicitly need the activation of a node itself besides the average of its neighbors. 
Four approximate algorithms are included for comparison: 
(1) NS, which adopts the NS estimator with no preprocessing. (2) NS+PP, which is same with NS but uses preprocessing. (3) CV+PP, which adopts the CV estimator and preprocessing. (4) CVD+PP, which uses the CVD estimator. All the four algorithms have similar low time complexity per epoch with $D^\bl=2$, while M1+PP takes $D^\bl=20$. We study how much convergence speed per epoch and model quality do these approximate algorithms sacrifice comparing with the M1+PP baseline.

We set the dropout rate as zero and plot the training loss with respect to number of epochs as Fig.~\ref{fig:iter-trainloss}. We can see that CV+PP can always reach the same training loss with M1+PP, while NS, NS+PP and IS+PP have higher training losses because of their biased gradients. CVD+PP is not included because it is the same with CV+PP when the dropout rate is zero. The results matches the conclusion of Theorem~\ref{thm:convergence}, which states that training with the CV estimator converges to a local optimum of Exact, regardless of $D^{(l)}$.

Next, we turn dropout on and compare the validating accuracy obtained by the model trained with different algorithms at each epoch. Regardless of the training algorithm, the exact algorithm is used for computing predictions on the validating set.
The result is shown in Fig.~\ref{fig:iter-acc}. We find that when dropout is present, CVD+PP is the only algorithm that can reach comparable validation accuracy with the exact algorithm on all datasets. Furthermore, its convergence speed with respect to the number of epochs is comparable with M1+PP, implying almost no loss of the convergence speed despite its $D^{(l)}$ is 10 times smaller. This is already the best we can expect - comparable time complexity with MLP, yet similar model quality with GCN. CVD+PP performs much better than M1+PP on the PubMed dataset, we suspect it finds a better local optimum.
Meanwhile, the simpler CV+PP also reaches a comparable accuracy with M1+PP for all datasets except PPI.
IS+PP works worse than NS+PP on the Reddit and PPI datasets, perhaps because sometimes nodes can have no neighbor selected, as we mentioned in Sec.~\ref{sec:importance-sampling}. Our accuracy result for IS+PP can match the result reported by~\citet{chen2018fastgcn}, while their NS baseline, GraphSAGE~\citep{hamilton2017inductive}, does not implement the preprocessing technique in Sec.~\ref{sec:preprocessing}.

\begin{table}
	\centering
\begin{tabular}{cccc}
	\toprule
	Alg.     & Valid.  acc. & Epochs & Time (s)  \\
	\midrule
	Exact & 96.0 & \textbf{4.8} & 252 \\
	NS & 94.4 & 102.0 & 445 \\
	NS+PP & 96.0 & 39.8 & 161\\
	IS+PP & 95.8 & 52.0 & 251 \\
	CV+PP & 96.0 & 7.6 & 39 \\
	CVD+PP & 96.0 & 6.8 & \textbf{37} \\
	\bottomrule
\end{tabular}
        \vspace{-3mm}
\caption{Time complexity comparison of different algorithms on the Reddit dataset.\label{tab:time-complexity}}
\end{table}

\begin{figure}
	\centering
\includegraphics[width=.85\linewidth]{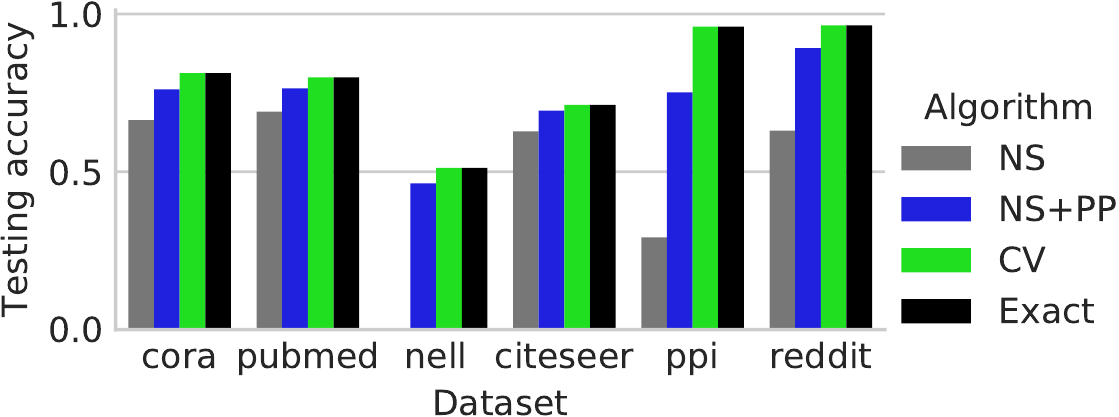}
\vspace{-4mm}
\caption{Comparison of the accuracy of different testing algorithms. The y-axis is Micro-F1 for PPI and accuracy otherwise.\label{fig:test}}
\end{figure}

\begin{figure}[t!]
	\centering\vspace{-.2cm}
	\includegraphics[width=0.85\linewidth]{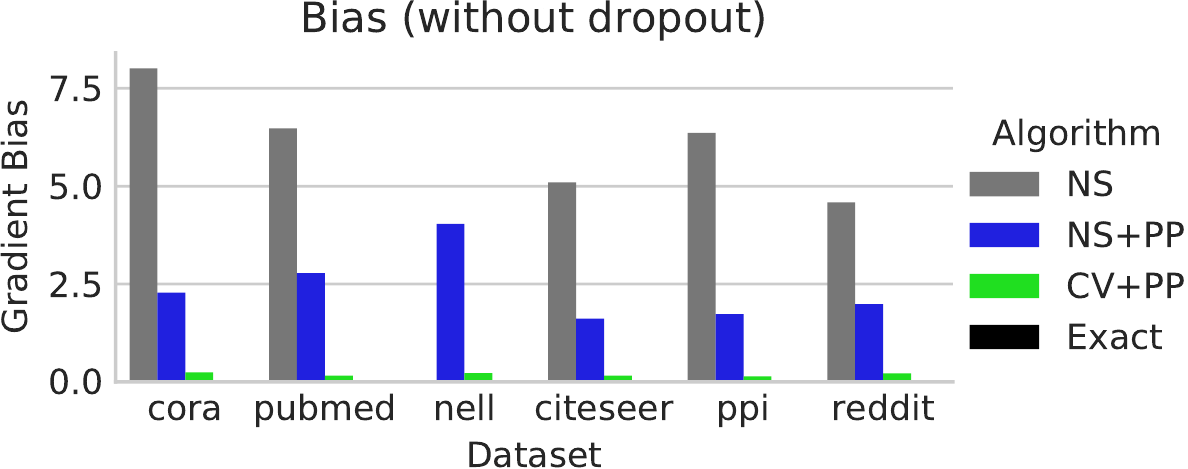}
	\includegraphics[width=0.85\linewidth]{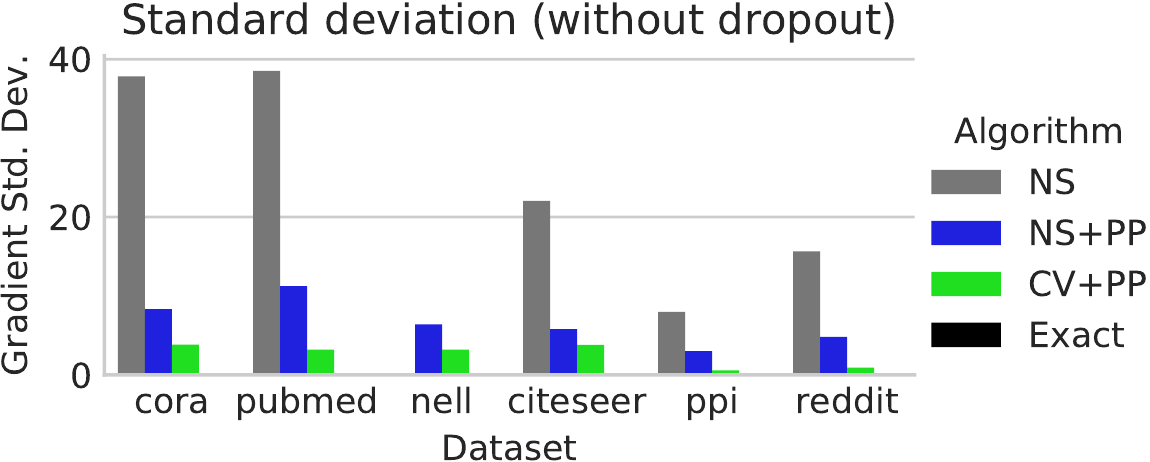}
	\includegraphics[width=0.85\linewidth]{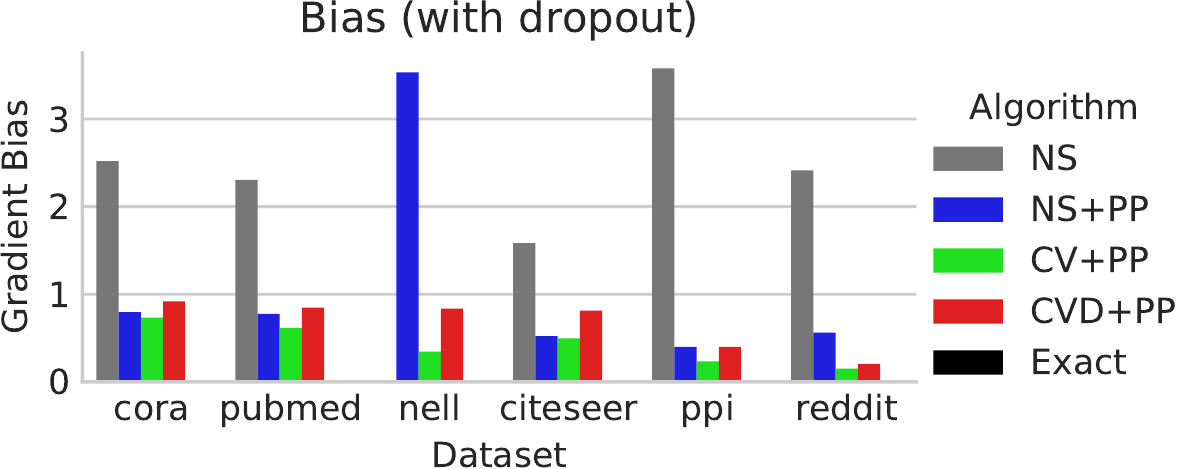}
	\includegraphics[width=0.85\linewidth]{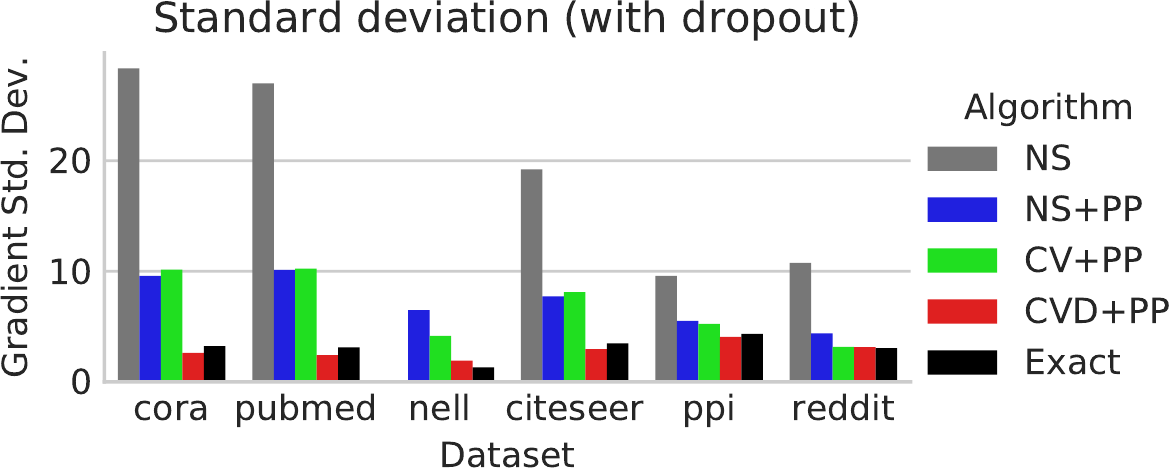}
	        \vspace{-4mm}
	\caption{Bias and standard deviation of the gradient for different algorithms during training. }
	\label{fig:var}
\end{figure}

\vspace{-.1cm}
\subsection{Further Analysis on Time Complexity, Testing Accuracy and Variance}\label{sec:experiment-testing}
\vspace{-.1cm}

Table~\ref{tab:time-complexity} reports the average number of epochs and time to reach a given 96\% validation accuracy on the largest Reddit dataset. Sparse and dense computations are defined in Sec.~\ref{sec:implementation}. We found that CVD+PP is about 7 times faster than M1+PP due to the significantly reduced receptive field size. Meanwhile, NS and IS+PP does not converge to the given accuracy.

We compare the quality of the predictions made by different algorithms, using the \emph{same} model  trained with M1+PP in Fig.~\ref{fig:test}.
As Theorem~\ref{thm:prediction} states, CV reaches the same testing accuracy as the exact algorithm, while NS and NS+PP perform much worse.

Finally, we compare the average bias and variance of the gradients  per dimension for first layer weights relative to the magnitude of the weights in Fig.~\ref{fig:var}.
For models without dropout, the gradient of CV+PP is almost unbiased. For models with dropout, the bias and variance of CV+PP and CVD+PP are usually smaller than NS and NS+PP.


\section{Conclusions}
The large receptive field size of GCN hinders its fast stochastic training. In this paper, we present control variate based algorithms to reduce the receptive field size. Our algorithms can achieve comparable convergence speed with the exact algorithm even the neighbor sampling size $D^{(l)}=2$, so that the per-epoch cost of training GCN is comparable with training MLPs. We also present strong theoretical guarantees, including exact prediction and the convergence to a local optimum.

\bibliography{sgd-gcn}
\bibliographystyle{icml2018}

\includepdfmerge{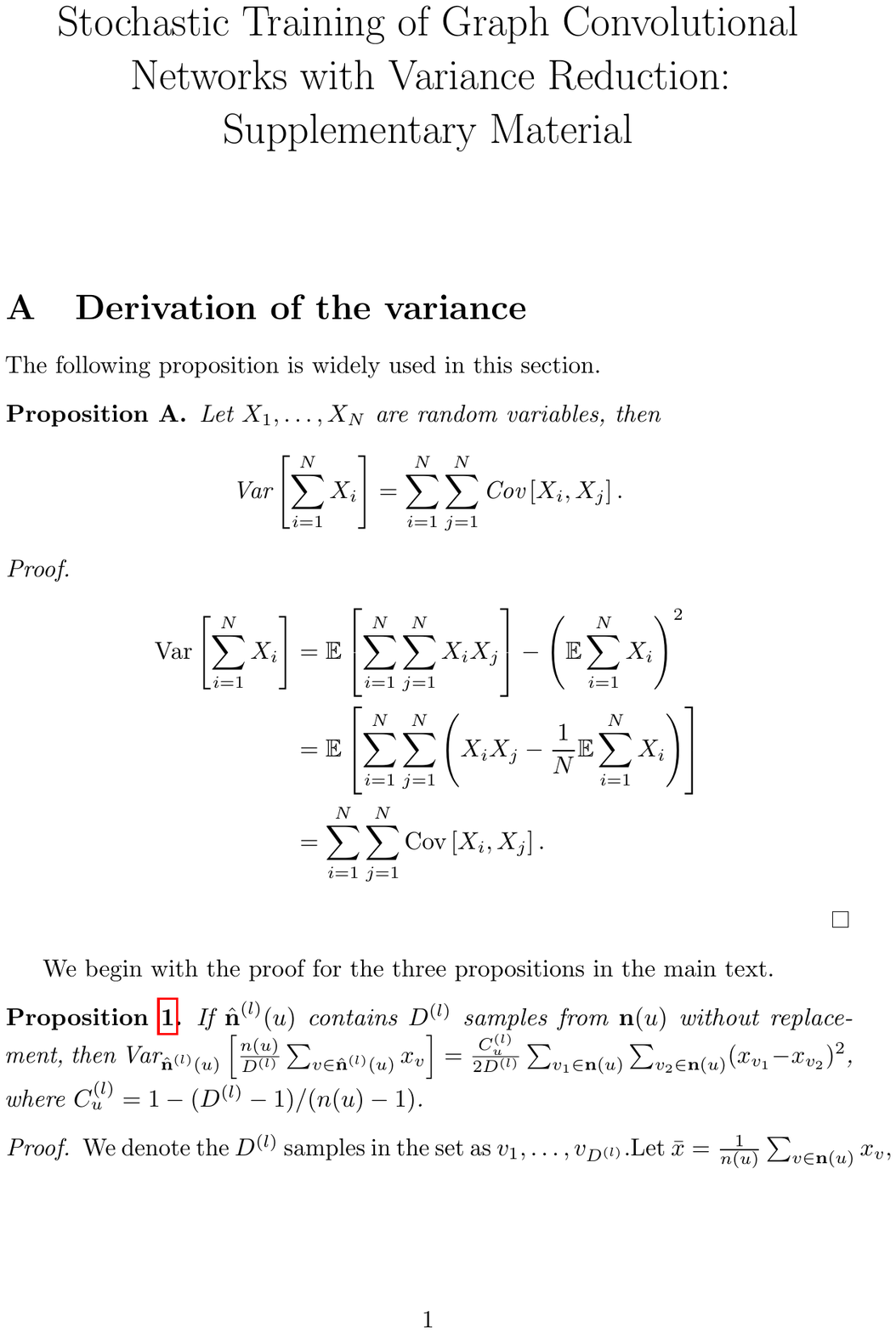,1-21}

\end{document}